\documentclass[11pt,a4paper]{article}
\usepackage[hyperref]{acl2021}
\usepackage{times}
\usepackage{latexsym}
\usepackage{graphicx}
\usepackage{booktabs}
\usepackage{microtype}

\aclfinalcopy 
\def\aclpaperid{3816} 

\usepackage{xcolor, soul}

\title{What Would a Teacher Do? Predicting Future Talk Moves}

\author{Ananya Ganesh, Martha Palmer \and Katharina Kann \\
  University of Colorado Boulder\\
  \texttt{\{ananya.ganesh, martha.palmer,  katharina.kann\}@colorado.edu}}

\date{}

\begin{document}
\maketitle
\begin{abstract}
Recent advances in natural language processing (NLP) have the ability to transform how classroom learning takes place. Combined with the increasing integration of technology in today's classrooms, NLP systems leveraging question answering and dialog processing techniques can serve as private tutors or participants in classroom discussions to increase student engagement and learning. 
To progress towards this goal, we use the classroom discourse framework of academically productive talk (APT) to learn strategies that make for the best learning experience. In this paper, we introduce a new task, called future talk move prediction (FTMP): it consists of predicting the next talk move -- an utterance strategy from APT -- given a conversation history with its corresponding talk moves. We further introduce a neural network model for this task, which outperforms multiple baselines by a large margin. Finally, we compare our model's performance on FTMP to human performance and show several similarities between the two.
\end{abstract}

\section{Introduction}
The field of natural language processing (NLP) has made rapid progress over the last few years \cite{superglue}. Success on natural language understanding, dialogue generation, and question answering tasks has spurred advances in NLP-based systems for educational applications.
\cite{mcnamara2013natural, litman2016natural, bea}.
Systems that can simulate human teachers in specific situations such as small-group discussions have the potential to aid learning by promoting student engagement. 
\begin{table}[ht]
    \centering \footnotesize
    \setlength{\tabcolsep}{3.pt}
    \begin{tabular}{p{7.5cm}}
 \toprule
 \textit{Scenario: Learning about proportional relationships in a classroom. The teacher gives an example of toasting two slices in a toaster, for 2 minutes.}\\
 \midrule
 \textbf{Teacher:} So we've just seen that 2 slices of toast gets done in 2 minutes. (\textit{None})\\
 \textbf{Teacher:} What if I had 3 slices of toast? (\textit{Press for Accuracy})\\
 \textbf{Student:} 4 minutes! (\textit{Wait}) \\
 \textbf{Teacher:} Why would it take 4 minutes? (\textit{Press for Reasoning}) \\
 \textbf{Student:} Because you'd have to use the toaster twice. (\textit{Wait})\\
 \midrule
 \textit{FTMP: Getting Students to Relate}\\ 
 \textit{~~~~~~~~~~~ (e.g., who else agrees it would be 4?)}\\
 \bottomrule
\end{tabular}
    \caption{Our proposed FTMP task; the teacher talk move corresponding to each utterance is shown in parentheses.}
    \label{tab:my_label}
\end{table}

Research has shown that deep conceptual learning is heightened when students are active participants in the classroom and contribute to discussions with their questions and ideas \cite{mcnamara2011measuring, bransford1999people}. However, large class sizes often make it difficult for all students to actively participate. Discussions in sub-groups increase each student's speaking time, but, in turn, make it impossible for a single teacher to guide all individual conversations. In this paper, we present a first step towards a system that can solve this problem by taking the teacher's role in facilitating sub-group discussions.

For this, we turn to a classroom discourse framework by \citet{michaels2008deliberative} called academically productive talk (APT). This framework, which we describe in detail in Section \ref{sec:apt}, provides both teachers and students with a set of \emph{talk moves} -- a family of utterance strategies to use for productive and respectful in-class discussions.  
As a first step towards developing an NLP system that can guide academically productive discussions, 
we aim to design a model which can predict
when which specific talk move is appropriate. Thus, we introduce the task of \textbf{future talk move prediction (FTMP)} -- given a conversation history,
the goal is to predict what the \textit{next} teacher talk move should be. We formulate this as a multi-class classification problem, with the input being a sequence of previous utterances and their corresponding talk moves, and the label being the next talk move. 

We further propose a model for FTMP, which we call 3-E.\footnote{Code for all models is available at \url{https://nala-cub.github.io/resources/}} It consists of three recurrent neural network (RNN) encoders: one for individual utterances, one for utterance sequences, and one for talk move sequences. The model is trained on transcripts of classroom discussions where teacher utterances have been annotated for the talk moves they represent. We consider the actions of the teacher to be our gold standard data for FTMP. We show that our model strongly outperforms multiple baselines and that adding sentence representations from RoBERTa \cite{Liu2019RoBERTaAR} or TOD-BERT \cite{wu-etal-2020-tod} -- a model trained on task-oriented dialogue -- does not increase performance further.

Finally, we investigate the performance of human annotators on FTMP. 
Unlike the teacher,
they do not have access to multi-modal signals, subject matter information, or knowledge of student behavior. This
setting, which mimics the
information available to our model,
is significantly different from the teachers who generate the gold standard utterances captured in our data.
We present a detailed analysis of their performance on a diagnostic test set, and highlight similarities to our model's performance. Our findings indicate that our model produces acceptable predictions a majority of the time. However, a gap between model and human performance on this task shows that there is still room for improvement.

\section{Academically Productive Talk}
\label{sec:apt}
\begin{table*}[ht]
    \centering
    \setlength{\tabcolsep}{3.pt}
    \begin{tabular}{p{3cm}|p{5.3cm}|p{6.8cm}}
 \toprule
 \textbf{Talk Move} & \textbf{Description} & \textbf{Example} \\ [0.5ex] 
 \midrule
 Keeping Everyone Together & Ask students to be active listeners & Raise your hand if you know the answer \\ 
 \midrule
 Getting Students to Relate & Ask students to contribute to another's ideas & Do you or agree or disagree with Michael? \\
 \midrule
 Revoicing & Repeat what a student says with adding words or rephrasing & S: It had two T: So it had two edges \\
 \midrule
 Restating & Repeat what a student says verbatim & S: Hexagon T: Hexagon! \\
 \midrule
 Press for Accuracy & Prompt for an answer & What is this called? \\
 \midrule
 Press for Reasoning & Prompt for explanation of thinking & How did you decide? \\
 \midrule
 \textit{None} & \textit{Fits into none of the above} & \textit{Good morning} \\ 
 \midrule
 \textit{Wait} & \textit{Teacher says nothing while student speaks} & \textit{S: It's the same shape} \\ [1ex]
 \bottomrule
\end{tabular}
    \caption{An overview of all teacher talk moves, their purpose and an example utterance. \emph{None} and \emph{Wait} are not APT talk moves, and represent generic utterances and teacher pauses during student utterances, respectively.}
    \label{tab:talkmoves}
\end{table*}

In this section, we provide an overview of the APT discourse framework and introduce a new task within the broader research area of NLP for educational applications: FTMP.

\subsection{Background on APT} 
Research in cognitive science and psychology highlights the importance of active participation as opposed to passive listening for achieving deep conceptual learning \cite{mcnamara2011measuring, bransford1999people, chi2014icap}. This can take the form of reflection on the lesson, as well as generation of new ideas, such as asking and answering questions, connecting concepts, and coming up with explanations. \citet{chapin2009classroom, goldenberg1992instructional, cazden1988classroom} discuss the importance of classroom conversations in this process. \citet{Chapin2003ClassroomDU} present case studies that show how implementing structured discussions in classrooms over a period of two years results in measurable improvements in test scores in mathematics.

To formalize how such discussions can be facilitated, \citet{michaels2008deliberative} present a classroom discourse framework called \emph{academically productive talk} (APT; also called \emph{accountable talk}). This includes strategies that teachers and students can use to promote engagement as well as deep conceptual learning through discussions.

\paragraph{Facets }
\citet{michaels2008deliberative} present three facets of accountability that APT encompasses: \emph{accountability to the learning community}, \emph{accountability to standards of reasoning}, and \emph{accountability to knowledge}. 
The first facet emphasizes the importance of listening to other students' contributions, and, subsequently, building on top of them. The second facet promotes talk that is based on evidence and reasoning, and involves getting students to provide explanations for their claims. The last facet covers talk which involves factual knowledge -- such as introducing a new concept, or challenging a student's claim to correct misconceptions.

\paragraph{Teacher Talk Moves} \citet{michaels2015conceptualizing} conceptualize the above facets as ``tools'' that can be used by teachers and students to engage in APT. For both teachers and students, these tools take the form of utterance strategies called \emph{talk moves}, which they can employ in order to conduct meaningful discussions. 

In this paper, we focus on the following six talk moves used by teachers: 
(1) \textit{\textbf{Keeping Everyone Together}} refers to utterances that manage student interactions, and asks students to be active listeners; 
(2) \textit{\textbf{Getting Students to Relate}} refers to utterances that ask a student to build on other students' ideas by agreeing, disagreeing, or following up; 
(3) \textit{\textbf{Restating}} occurs when a teacher repeats a student's answer or claim verbatim with the purpose of ensuring it reaches the entire classroom; 
(4) \textit{\textbf{Revoicing}} happens when a teacher paraphrases a student's ideas, but adds or removes information in order to correct a student or convey new knowledge; 
(5) \textit{\textbf{Pressing for Reasoning}} refers to utterances that ask a student to explain a decision or to connect multiple ideas; and
(6) \textit{\textbf{Pressing for Accuracy}} refers to utterances that prompt for answers to a factual question, e.g., about a method or a result.

\textit{Keeping Everyone Together, Getting Students to Relate}, and \textit{Restating} are part of accountability to the learning community; \textit{Revoicing} and \textit{Press for Reasoning} are part of accountability to standards of reasoning, and \textit{Press for Accuracy} falls under accountability to knowledge. Examples for all talk moves are shown in Table \ref{tab:talkmoves}.

\paragraph{Student Talk Moves} While we do not focus on student talk moves in this work, we summarize them here for completeness. Student talk moves can also be grouped into the same accountability facets as teacher talk moves \cite{o2019supporting}. Under accountability to the learning community, we have \textit{Relating to Another Student} -- building on a classmate's ideas or asking questions about them, and \textit{Asking for more info} -- requesting help from the teacher on a problem. Under accountability to knowledge, there is \textit{Making a Claim} -- providing an answer or a factual statement about a topic. Under accountability to standards of reasoning, we have \textit{Providing Evidence/Explanation} -- explaining their thinking with evidence.

\subsection{Future Talk Move Prediction}
In order to build a system that can facilitate in-class discussion in the way a human would, we aim at automatically answering the question \emph{What would a teacher do?} at each point within a classroom conversation. Specifically, we define the task of future talk move prediction (FTMP) as choosing the next appropriate teacher talk move to make, given the history of what has been discussed so far. 

Formally, the input for FTMP is a dialogue context $C = c_0, c_1, ..., c_{t}$, with each context element consisting of an utterance $u_i$, a binary variable  $s_i$ indicating if the speaker is different from the previous utterance, and a teacher talk move label $t_i$, i.e., $c_i = (s_i, u_i, t_i)$.
The goal then is to predict the next teacher talk move $t_{t+1}$ out of the possible talk moves defined above. Note that the future utterances are unseen; the prediction of the next talk move is to be made only based on the conversation history.

\section{Related Work}
\subsection{Promoting APT with NLP Systems}
Ideas from APT have been incorporated with success into intelligent tutoring systems \cite{dyke2013towards, tegos2016conversational, adamson2014towards}. These systems provide an environment to simulate classroom discussions, for instance, as small groups collaboratively solving problems with a shared textual chat interface for communication. The intelligent agent then plays a role similar to a teacher -- it monitors the conversations and makes decisions about when to intervene in order to promote student engagement and learning.

\citet{adamson2014towards} study the effects of two specific interventions: using the \textit{Revoicing} talk move as well as an \textit{Agree/Disagree} talk move (which corresponds to the \textit{Getting Students to Relate} talk move in our above categorization). These interventions are made by matching student utterances in the chat to an annotated set of concepts and misconceptions for the topic being taught. 
Through multiple case studies, they show that interventions by the agent have a positive effect on learning, as measured by test scores before and after using the system. The agent interventions also prove useful in increasing student talk frequency. Similarly, \citet{tegos2015promoting} find that an APT-based intervention called Linking Contributions, similar to \textit{Getting Students to Relate}, improves explicit reasoning as well as learning outcomes in students.

The findings of the above work provide a strong motivation for building a conversational AI system that can produce academically productive talk. Unlike the above systems, which focused particularly on accountability to the learning community, we attempt to predict opportunities for intervention across all talk moves described in Section \ref{sec:apt}. Since we do not have access to gold annotations of statements corresponding to concepts and misconceptions, we make use of transcripts of classroom discourse with annotations for talk moves used by the teacher. 

\subsection{NLP for Educational Applications}
Our work is a first step towards improving in-class discussions with the help of an NLP system and, thus, to improve student learning and engagement. Prior work in understanding classroom discourse using NLP includes \citet{suresh2019automating} and \citet{donnelly}. They propose an application where feedback can be provided to teachers by automatically classifying their utterances into talk moves. Other applications of NLP to education include language learning assistance \cite{beatty2013teaching, carlini2014improving, tetreault2014bucking}, writing assistance \cite{dahlmeier2011correcting, chollampatt2018multilayer, chukharev2016causal}, and automated scoring \cite{burstein1998automated, farag2018neural, beigman-klebanov-madnani-2020-automated}.

\subsection{Dialogue Systems}
Our work is further related to research on dialogue systems. Similar to talk moves, dialogue acts provide a categorization for utterances, but, in contrast to talk moves, they apply to general-purpose conversations \cite{stolcke2000dialogue, calhoun2010nxt}. Examples include \textit{Statement, Question, Greeting}, and \textit{Apology}. Dialogue act tagging, which is sometimes called dialogue act prediction, is the task of classifying an utterance into the category it belongs to \cite{yudialog, khanpour-etal-2016-dialogue, wu-etal-2020-tod}. Analogous to FTMP, future dialogue act prediction is the task of predicting what the next dialogue act should be, given a conversation history \cite{tanaka-etal-2019-dialogue}.

Pretrained models have been successfully adapted to the task of dialogue generation \cite{zhang-etal-2020-dialogpt, wu-etal-2020-tod, adiwardana2020humanlike, roller2020recipes}. However, if directly used in the classroom, these models could potentially produce harmful or unsuitable dialogue as they are trained on large datasets comprising conversations from the internet \cite{bender2021dangers}. Additionally, we want a system to facilitate structured conversations, and not cause further diversions -- this is in contrast to many task-oriented or open-domain dialogue systems whose purpose is to entertain and appear personable to the user. Hence, we propose FTMP as a crucial first step towards an NLP system capable of facilitating classroom discussions.

\begin{figure*}[t]
    \centering
    \includegraphics[angle=270, clip, trim=2cm 0.5cm 2cm 0.5cm, width=.7\textwidth]{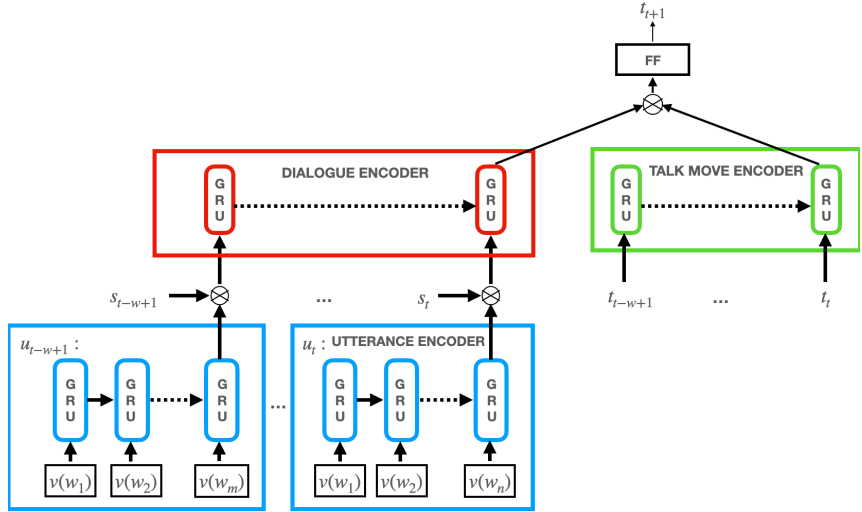}
    \caption{Our proposed model for FTMP, consisting of separate encoders for utterances, past task moves, and the overall context.}
    \label{fig:model}
\end{figure*}
\section{Model}
\label{sec:model}
In this section, we describe our proposed model for FTMP, cf. Figure \ref{fig:model}. Following \citet{tanaka-etal-2019-dialogue}'s model for future dialogue act prediction, its main components are three encoders. We hence name our model \textbf{3-E}.
Our model predicts the next teacher talk move $t_{t+1}$, given the last $w$ context elements $c_{t-w+1}, \dots, c_t$.

\paragraph{Utterance Encoder } The first encoder -- the utterance encoder -- is a single-layer gated recurrent unit \citep[GRU;][]{cho-etal-2014-learning}. It processes the sequence of vector representations $v(w_1), \dots, v(w_m)$ of the words $w_1, \dots, w_m$ that each utterance $u_i$ consists of and computes the last hidden state as a vector representation of $u_i$:
\begin{equation}
    \hat{a}_i = \textrm{GRU}(v(w_1), \dots, v(w_m))
\end{equation}
Each utterance representation is then concatenated with a representation $s_i$ of the speaker role. This representation is either 1 or 0, depending on if the speaker has changed from the previous utterance: 
\begin{equation}\label{eq:utt}
    a_i = \textrm{cat}(\hat{a}_i, s_i)
\end{equation}

\paragraph{Dialogue Encoder } Next, the sequence of all $w$ utterance representations is passed to the dialogue encoder, which is also a single-layer GRU. The dialogue encoder processes the sequence, and we take the last hidden state as a representation of all utterances within our context window:
\begin{equation}
    b_t = \textrm{GRU}(a_{t-w+1}, \dots, a_t)
\end{equation}

\paragraph{Talk Move Encoder } The talk move encoder is a third single-layer GRU, which encodes the sequence of vector representations $v(t_{t-w+1}), \dots, v(t_{t})$ of talk moves $t_{t-w+1}, \dots, t_{t}$:
\begin{equation}
    d_t = \textrm{GRU}(v(t_{t-w+1}), \dots, v(t_{t}))
\end{equation}

We obtain our final context representation $r_t$ by concatenating the representation of all utterances and all talk moves within the context window:
\begin{equation}\label{eq:concattm}
    r_t = \textrm{cat}(b_t, d_t)
\end{equation}

Finally, we pass $r_t$ through a two-layer feed-forward network and a softmax layer to obtain a probability distribution over possible future talk moves.

\subsection{Adding a Pretrained Sentence Encoder}
\paragraph{RoBERTa } Pretrained models  define the state of the art on a large variety of NLP tasks \cite{superglue}. Thus, we additionally experiment with concatenating an utterance representation 
computed by RoBERTa \cite{Liu2019RoBERTaAR} to the output of 3-E's utterance encoder. Equation \ref{eq:utt} then becomes:
\begin{equation}\label{eq:roberta}
    a^*_i = \textrm{cat}(\hat{a}_i, s_i, \textrm{RoBERTa}(w_1, \dots, w_m))
\end{equation}
We call the model with additional RoBERTa representations \textbf{3-E-RoBERTa}.

\paragraph{TOD-BERT } Since there is a  domain mismatch between the text that RoBERTa is trained on and our data, we further experiment with including a model trained on task-oriented dialogue, called TOD-BERT \cite{wu-etal-2020-tod}. TOD-BERT differentiates between user utterances and system utterances using two special tokens, [USR] and [SYS]. Correspondingly, we use the [USR] token to indicate student utterances and the [SYS] token to indicate teacher utterances. We then concatenate a context of $w$ utterances, marked by speaker tokens when there is a change in speaker, to obtain $c_{tod}$. Finally, we encode $c_{tod}$ using the pretrained TOD-BERT model and concatenate it with the output of the dialogue encoder and talk move encoder. Equation \ref{eq:concattm} then becomes:
\begin{equation}\label{eq:tod}
    r_t = \textrm{cat}(b_t, d_t, \textrm{TOD-BERT}(c_{tod}))
\end{equation}
We call this model \textbf{3-E-TOD-BERT}. When pretrained sentence encoders are used, we use the respective BPE \cite{sennrich-tokenize} tokenizer for each model.

\subsection{Computing the Loss}
\label{sec:loss}
We train 
all models using a cross-entropy loss. However, we observe a strong class imbalance in our training data, cf. Figure \ref{fig:tm_distribution}.
Thus, we compute label weights inversely proportional to the frequency of a label's occurrence in the data and use them to weight the loss for each training example.

\section{Experiments}
\subsection{Dataset}
For our experiments, we make use of the dataset from
\citet{suresh2019automating}. It consists of 216 annotated transcripts of classroom discourse collected in public schools in the US. The topic of instruction is mathematics. The transcripts have been collected from classes from kindergarten to grade 12 and are all in English. Each row in the transcripts consists of an utterance, the name of the speaker, and the talk move realized by this utterance. 

The annotations assign each teacher utterance to one of the 6 APT talk moves described in Section \ref{sec:apt}. Utterances that do not fit into any talk move category are coded as \textit{None}. In addition, we designate the teacher talk move corresponding to utterances made by a student as \textit{Wait}. 
This category is needed as we eventually want to be able to detect when an in-class NLP system should remain quiet.
The original annotations contain two additional categories that we remove due to sparsity: $Marking$ refers to repeated utterances, and we merge it with the $Restating$ category. Some student utterances are annotated as $Context$, which we merge with the $Wait$ category.

\begin{figure}
    \centering
    \includegraphics[scale=0.9, clip, trim=0cm 0.3cm 0cm 0cm, width=\columnwidth]{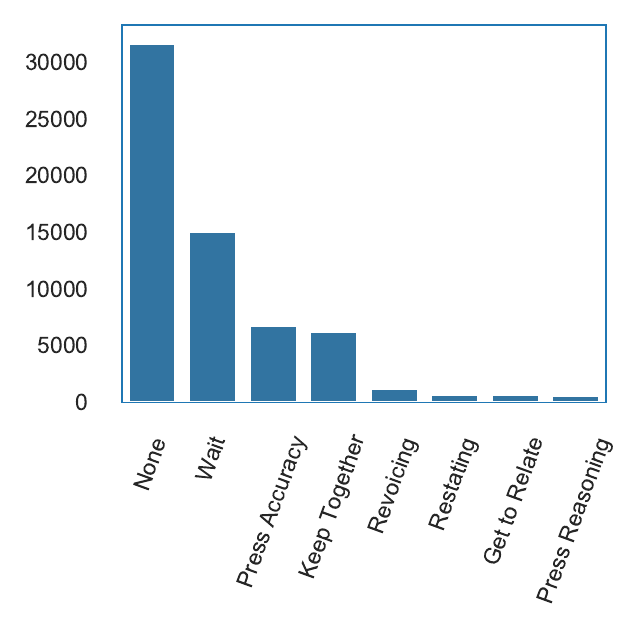}
    \caption{Distribution of talk moves in our training set.}
    \label{fig:tm_distribution}
\end{figure}

We create training, development and test data from $70\%$, $15\%$, and $15\%$ of the available documents, respectively. Thus, we have 151 documents for training, and 32 documents for each of development and testing. 
Our training set consists of over 63k utterances, and the distribution of talk moves in the training set is shown in Figure \ref{fig:tm_distribution}.

Since 3-E's utterance encoder operates on the word level, we split each utterance into words using the NLTK word tokenizer \cite{nltk}.

\subsection{Baselines}
We compare our model to three baselines. 

\paragraph{Random Baseline (RB)} This baseline randomly selects one of the 8 talk moves for each input.

\paragraph{Talk Move Bigram Model (TMBM)} For this baseline, we compute the conditional probability of every talk move in the training set, given the talk move realized by the previous utterance. We then pick the talk move with the highest conditional probability. 

\paragraph{Talk Moves Only (TM-only)} We further train a GRU model exclusively on the sequences of prior talk moves, i.e., this baseline has no access to actual utterances. We implement two variants of this baseline, one with class weights for training (TM-only-w), and one without (TM-only-z).

\subsection{Metrics} Since the classes in our dataset are highly imbalanced, we do not evaluate using accuracy. Instead, we report precision, recall, and F1 score for all models. We compute  F1 for all 8 classes individually, and additionally calculate macro-average F1 as an overall score for our dataset. 

\subsection{Results}
\begin{table*}[t]
    \centering
    \footnotesize
    \setlength{\tabcolsep}{3.pt}
    \begin{tabular}{l||r|r|r||p{0.7cm}|p{0.7cm}|p{0.8cm}|p{1.2cm}|p{1.3cm}|p{1.2cm}|p{1.3cm}|p{1cm}}
 \toprule
 \textbf{Model} & \textbf{Prec.} & \textbf{Recall} & \textbf{F1} & \textbf{None F1} & \textbf{Wait F1} & \textbf{Press Acc. F1} & \textbf{Keep Together F1} & \textbf{Revoicing F1} & \textbf{Getting Students to Relate F1} & \textbf{Restating F1} & \textbf{Press Reasoning F1}\\ [0.5ex] 
 \midrule
 3-E & 35.67 & 30.38 & \textbf{29.84} & 72.72 & 75.70 & 24.25 & 13.31 & 20.27 & \textbf{3.25} & \textbf{18.45} & \textbf{10.77}\\
 \midrule
 3-E-ToD-BERT & 31.10 & 28.92 & 28.51 & \textbf{73.05} & \textbf{77.67} & \textbf{25.18} & 13.81 & 18.89 & 0.00 & 17.92 & 1.53 \\
 \midrule
 3-E-RoBERTa & 33.81 & 28.04 & 27.62 & 69.89 & 73.34 & 20.48 & \textbf{14.68} & 20.31 & 1.57 & 17.81 & 2.88 \\
 \midrule
 TM-only-w & 28.95 & 22.66 & 20.40 & 72.14 & 52.75 & 13.19 & 1.94 & \textbf{21.04} & 0.68 & 0.00 & 1.45 \\
 \midrule
 TM-only-z & 18.38 & 18.81 & 16.43 & 71.60 & 52.14 & 0.22 & 0.00 & 7.47 & 0.00 & 0.00 & 0.00\\
 \midrule
 RB & 12.25 & 11.74 & 8.50 & 15.50 & 19.82 & 10.52 & 2.93 & 2.28 & 2.81 & 11.58 & 2.57 \\ 
 \midrule
 Majority & 6.46 & 12.50 & 8.52 & 68.16 & 0.00 & 0.00 & 0.00 & 0.00 & 0.00 & 0.00 & 0.00 \\
 \midrule
 TMBM & 13.18 & 17.74 & 15.09 & 49.59 & 71.11 & 0.00 & 0.00 & 0.00 & 0.00 & 0.00 & 0.00\\ 
 \midrule
 \midrule
 \textit{Annotator 1} & \textit{37.57} & \textit{29.29} & \textit{24.50} & \textit{31.46} & \textit{42.55} & \textit{15.69} & \textit{18.18} & \textit{29.51} & \textit{5.13} & \textit{53.52} & \textit{0.00} \\
 \midrule
 \textit{Annotator 2} & \textit{38.99} & \textit{33.16} & \textit{30.51} & \textit{25.19} & \textit{54.17} & \textit{36.17} & \textit{20.00} & \textit{24.66} & \textit{21.74} & \textit{42.62} & \textit{9.52} \\
 \midrule
 \textit{3-E} & \textit{15.83} & \textit{20.22} & \textit{14.57} & \textit{29.32} & \textit{26.53} & \textit{4.44} & \textit{12.66} & \textit{8.16} & \textit{0} & \textit{35.48} & \textit{0.00} \\
 \bottomrule
\end{tabular}
\caption{Model and annotator performance on FTMP. Italics indicate results on a diagnostic test set of 300 examples taken from the development set.}
\label{tab:result1}
\end{table*}

Table 2 shows the performance of our proposed model 3-E as well as of 3-E-TOD-BERT, 3-E-RoBERTa, and all baselines.
Looking at the macro-average F1 scores, we see that 3-E performs best with an F1 of 29.84. 
3-E-RoBERTa, with 27.62 F1, performs worse; however, given that this model has more parameters and includes a strong pretrained component, this is an unexpected result.\footnote{We further experiment with directly finetuning RoBERTa on our task, but find its performance to be poor overall (around 17 F1). Hence, we do not report detailed results.}

To reduce the domain mismatch between RoBERTa's training data and our classroom dialogue data, we substitute RoBERTa with TOD-BERT, which is also trained on dialogue. We see that, while 3-E-TOD-BERT performs better than 3-E-RoBERTa, 3-E still outperforms it. We also observe that on most individual talk move classes, 3-E-TOD-BERT performs equal to or better than 3-E. However, it does poorly on a few classes that are less prevalent in the data. We hypothesize that small changes in the quality of the utterance representations have negligible effect on our model, since it gets a large amount of information from the sequence of prior talk move labels.
This hypothesis is supported by the fact that all baselines which only receive prior talk move labels as input, i.e., TM-only-w, TM-only-z and TMBM, obtain F1 scores of 20.40, 16.42, and 15.09, respectively. All of them strongly outperform a random baseline with an F1 of 8.50. Comparing 3-E to our baselines, we see that our proposed model is indeed strong on FTMP: 3-E outperforms the best baseline, TM-only-w, by 9.44 F1.

\section{Analysis}
\begin{figure*}[th]
    \centering
    \includegraphics[width=2\columnwidth, height=5.7cm, clip, trim=0.25cm 0.1cm 0.2cm 0.3cm,]{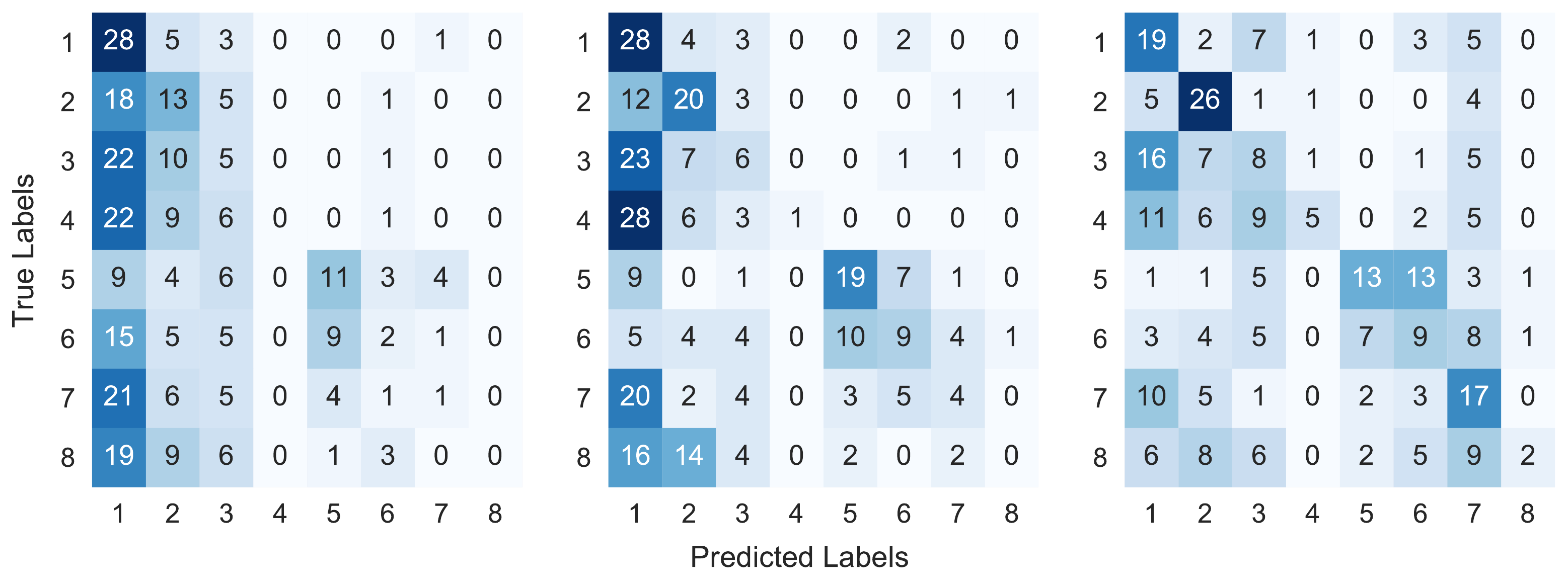}
    \caption{Confusion matrix on the diagnostic test set for 3-E (left) and our two annotators (middle and right). The talk move labels in order are: \textit{None, Wait, Press for Accuracy, Keeping Everyone Together, Revoicing, Getting Students to Relate, Restating}, and \textit{Press for Reasoning}.}
    \label{fig:confusion}
\end{figure*}

\begin{table}[t!]
    \centering
    \footnotesize
    \setlength{\tabcolsep}{8.5pt}
    \begin{tabular}{l|r|r|r}
 \toprule
 \textbf{Label} & \textbf{Prec.} & \textbf{Recall} & \textbf{F1}\\
 \midrule
 Macro average & 43.76 & 42.65 & 42.44\\
 \midrule
 None & 68.30 & 77.76 & 72.72\\
 Wait & 71.48 & 80.45 & 75.70\\
 
 Learning Community & 29.59 & 14.10 & 19.10 \\
 Content Knowledge & 27.86 & 21.47 & 24.25\\
 Rigorous Thinking & 21.56 & 19.44 & 20.45\\
 \bottomrule
\end{tabular}
\caption{Performance of 3-E evaluated on facets.}
\label{tab:result_facets}
\end{table}

\subsection{FTMP on the Facet Level}
In some cases, the distinctions between different talk moves are subtle. For instance, both the \textit{Keeping Everyone Together} and \textit{Getting Students to Relate} moves, which fall under the facet of accountability to the learning community, are made when the teacher wants the students to actively listen and respond to statements made in the classroom. To understand how well the model can distinguish between different accountability facets, we evaluate our best model, 3-E, on the facet level by binning all predicted talk moves into their corresponding facets for the computation of the F1 score.

In Table \ref{tab:result_facets}, we see that performance goes up by 12.60 points in this setting, indicating that 3-E is able to distinguish between labels at a coarse-grained level, but struggles with fine-grained distinctions.

\subsection{Window Size and Class Weights}
We further investigate the effect of weighting the loss as described in Section \ref{sec:loss}
and the influence of different context window sizes. Full results on the development set can be found in Table \ref{tab:tuning} in the appendix, and we provide a summary of our findings here.

Varying the window size leads to small changes in F1. For smaller window sizes of 1 and 2, F1 is slightly lower at 28.64 and 27.62. When the window size is increased to 5, F1 increases to 29.83. However, when the window size is increased further to 7, F1 drops slightly, to 29.05. We thus choose a window size of 5 to train all our models, and conclude that very large window sizes are not  beneficial. We hypothesize that this might be due to the most relevant information for FTMP being contained in the most recent dialogue history.   

Further, class weighting during training increases 3-E's F1 score 
by 3, from 26.94 to 29.83. We thus, conclude that class weights are important to account for the label imbalance in our training set.
\footnote{We also experiment with downsampling the dominant classes and find its performance to be comparable to class weighting.}

\begin{table}[t]
    \centering
    \footnotesize
    \setlength{\tabcolsep}{7.pt}
    \begin{tabular}{l|r}
    \toprule
      \multicolumn{1}{l}{\textit{Primary option: Annotators}} \\
      \midrule
      Inter-annotator agreement & 46\% \\
      Annotator 1--ground truth agreement & 29\% \\
      Annotator 2--ground truth agreement & 33\% \\
      Both Annotators--ground truth agreement & 17\% \\
      \midrule 
      \multicolumn{1}{l}{\textit{Primary option: Model}} \\
      \midrule
      Model--Annotator 1 agreement & 48\% \\
      Model--Annotator 2 agreement & 33\% \\
      Model--ground truth & 20\% \\
      \midrule
      \multicolumn{1}{l}{\textit{Acceptable options}} \\
      \midrule
      Annotator 1's primary accepted by Annotator 2 & 94\% \\
      Annotator 2's primary accepted by Annotator 1 & 91\% \\
      Ground truth accepted by Annotator 1 & 72\% \\
      Ground truth accepted by Annotator 2 & 79\% \\
      Model predictions accepted by Annotator 1 & 90\% \\
      Model predictions accepted by Annotator 2 & 84\% \\
      \bottomrule
    \end{tabular}
    \caption{Percentage agreement between our annotators, the ground truth, and the model's predictions on the diagnostic test set.}
    \label{tab:humananno}
\end{table}
\subsection{Performance of Human Annotators}
We further investigate (1) the difficulty of FTMP for human annotators, (2) the effect of multiple choices for the future talk move as opposed to a single answer, and (3) how annotator decisions differ from 3-E's predictions. While our gold standard are actions of a teacher, who also represent human performance, an FTMP annotator is different from a teacher since the former is presented with the exact same information as our model. In contrast, we expect a teacher's decisions to be informed by background knowledge about the students, knowledge about the content being discussed, and  multi-modal information.

We recruit two annotators who have extensive experience with linguistic annotation tasks, and are familiar with talk moves. We present them with a diagnostic test set of 300 examples from the development set. Similar to the model input, each example consists of the past 5 utterances,  the corresponding talk moves, and speaker information. Both annotators then provide (1) the most likely future talk move, referred to as the `primary' option, and (2) a set of all acceptable future talk moves given the conversation history. As with our modeling setup, we consider the ground truth for the primary option to be the talk move made by the teacher in the classroom transcript. Each talk move is equally distributed in the ground truth, with 37 examples each of talk moves \textit{None, Wait, Restating, Revoicing}, and 38 examples each of the other talk moves. 

\paragraph{Primary Option } The last 3 rows in Table \ref{tab:result1} show the performance of our annotators' primary option and the model on the diagnostic test set. There is a significant gap of 10 F1 and 15 F1 respectively between the performance of the model and the two annotators. However, there are similarities in the class-wise breakdown. Both the annotators and the model achieve a high F1 on the classes \textit{None, Wait,} and \textit{Restating}, and perform poorly on \textit{Press for Reasoning} and \textit{Getting Students to Relate}.  \textit{Press for Reasoning}, and \textit{Getting Students to Relate} are the least prevalent classes in the data. However, the annotators' performance on these
talk moves 
suggests that these classes are intrinsically more difficult to predict based on conversational cues alone. On the other hand, the model's poor performance on categories like \textit{Revoicing} and \textit{Keeping Everyone Together} in comparison to the annotators indicates that 
there is still room for improvement for our model.

The similarities between the model's predictions and the annotators' primary option is further illustrated by the confusion matrices in Figure  \ref{fig:confusion}. Both the model as well as our annotators erroneously predict \textit{None} when the true label is another category, both confuse \textit{Restating} and \textit{Revoicing} for each other, and both erroneously predict \textit{Keeping Everyone Together} when the true category is \textit{Getting Students to Relate} or \textit{Press for Reasoning}. 

\paragraph{Acceptable Options } Table \ref{tab:humananno} shows the percentage of responses for which the annotators agree with each other, the ground truth, and the model's predictions. On average, both annotators provide 3 acceptable options in addition to the primary -- thus, roughly half the classes were viewed as acceptable for most examples. The impact of having a set of acceptable options in addition to a single correct option is evident here: while inter-annotator agreement is only around 46\% on the primary option alone, the primary option of each annotator was one of the acceptable options by the other annotator in over 90\% of the cases. Additionally, while agreement between the ground truth and the primary option is low with 29\% and 33\%, this increases to 72\% and 79\% when additional options are being considered. 

Table \ref{tab:humananno} helps us contextualize our model's performance. Interestingly, the annotators agree with the predictions made by the model more often than they agree with the ground truth. This indicates that the model might truly be grasping overall patterns and cues from the training data, but probably struggles with finer-grained distinctions between the talk move classes. This is further substantiated by our analysis of how often the model's predictions featured in the set of acceptable options for each annotator. We find that the predictions were acceptable in 90\% and 84\% of all instances respectively for each annotator.

\section{Conclusion}
In this paper, we made use of the APT discourse framework to take a first step towards a system that can fill the role of a teacher in classroom discussions. 
We introduced the task of FTMP, which consists of predicting the next appropriate talk move given an in-class dialogue context. We then presented 3-E, a model for the task, which outperforms multiple baselines. Finally, we conducted an analysis of human performance on FTMP, and compared it to our model. Our results showed that, while the task is challenging, our model produces acceptable talk moves and can identify overall patterns, indicated by similarities with human performance.

\section*{Acknowledgments}
We would like to thank the anonymous reviewers for their insightful feedback. We would also like to thank the researchers involved in the NSF TalkBack project at the University of Colorado for giving us access to this data, and for their assistance with this work. We would also like to thank Amanda Howard, Charis Clevenger and Jennifer Jacobs for their feedback and assistance with the human annotations, and the members of the NALA research group for their valuable suggestions. This research was supported by the NSF National AI Institute for Student-AI Teaming (iSAT) under grant DRL 2019805. The opinions expressed are those of the authors and do not represent views of the NSF.

\section*{Ethics and Impact Statement}
The data used in this paper was collected after obtaining informed consent from all participants, and this research was approved by the University of Colorado Boulder’s Institutional Review Board (protocol \#18-0432). All authors completed appropriate training prior to accessing the data. Since this dataset was provided to us by its owners purely for the purpose of the research reported in this paper, we are not making it publicly available here.

Further, this research is part of the NSF National AI Institute for Student-AI Teaming. Several learning scientists and educators are participating in the institute and advising us on avoiding unintended consequences and harms that could stem from NLP research. The work described in this paper is preliminary, and will be used to inform strategies for additional data collection and model design. Field trials in carefully controlled classroom settings will be conducted before wider deployment.

\bibliographystyle{acl_natbib}
\bibliography{anthology,acl2021}

\begin{thebibliography}{46}
\expandafter\ifx\csname natexlab\endcsname\relax\def\natexlab#1{#1}\fi

\bibitem[{Adamson et~al.(2014)Adamson, Dyke, Jang, and
  Ros{\'e}}]{adamson2014towards}
David Adamson, Gregory Dyke, Hyeju Jang, and Carolyn~Penstein Ros{\'e}. 2014.
\newblock Towards an agile approach to adapting dynamic collaboration support
  to student needs.
\newblock \emph{International Journal of Artificial Intelligence in Education},
  24(1):92--124.

\bibitem[{Adiwardana et~al.(2020)Adiwardana, Luong, So, Hall, Fiedel,
  Thoppilan, Yang, Kulshreshtha, Nemade, Lu, and Le}]{adiwardana2020humanlike}
Daniel Adiwardana, Minh-Thang Luong, David~R. So, Jamie Hall, Noah Fiedel,
  Romal Thoppilan, Zi~Yang, Apoorv Kulshreshtha, Gaurav Nemade, Yifeng Lu, and
  Quoc~V. Le. 2020.
\newblock \href {http://arxiv.org/abs/2001.09977} {Towards a human-like
  open-domain chatbot}.

\bibitem[{Beatty(2013)}]{beatty2013teaching}
Ken Beatty. 2013.
\newblock \emph{Teaching \& researching: Computer-assisted language learning}.
\newblock Routledge.

\bibitem[{Beigman~Klebanov and
  Madnani(2020)}]{beigman-klebanov-madnani-2020-automated}
Beata Beigman~Klebanov and Nitin Madnani. 2020.
\newblock \href {https://doi.org/10.18653/v1/2020.acl-main.697} {Automated
  evaluation of writing {--} 50 years and counting}.
\newblock In \emph{Proceedings of the 58th Annual Meeting of the Association
  for Computational Linguistics}, pages 7796--7810, Online. Association for
  Computational Linguistics.

\bibitem[{Bender et~al.(2021)Bender, Gebru, McMillan-Major, and
  Shmitchell}]{bender2021dangers}
Emily~M Bender, Timnit Gebru, Angelina McMillan-Major, and Shmargaret
  Shmitchell. 2021.
\newblock On the dangers of stochastic parrots: Can language models be too big.

\bibitem[{Bransford et~al.(1999)Bransford, Brown, and
  Cocking}]{bransford1999people}
John~D Bransford, Ann~L Brown, and Rodney~R Cocking. 1999.
\newblock \emph{How people learn: Brain, mind, experience, and school}.
\newblock National Academies Press.

\bibitem[{Burstein et~al.(2020)Burstein, Kochmar, Leacock, Madnani, Pil{\'a}n,
  Yannakoudakis, and Zesch}]{bea}
Jill Burstein, Ekaterina Kochmar, Claudia Leacock, Nitin Madnani, Ildik{\'o}
  Pil{\'a}n, Helen Yannakoudakis, and Torsten Zesch. 2020.
\newblock Proceedings of the fifteenth workshop on innovative use of nlp for
  building educational applications.
\newblock In \emph{Proceedings of the Fifteenth Workshop on Innovative Use of
  NLP for Building Educational Applications}.

\bibitem[{Burstein et~al.(1998)Burstein, Kukich, Wolff, Lu, Chodorow,
  Braden-Harder, and Harris}]{burstein1998automated}
Jill Burstein, Karen Kukich, Susanne Wolff, Chi Lu, Martin Chodorow, Lisa
  Braden-Harder, and Mary~Dee Harris. 1998.
\newblock Automated scoring using a hybrid feature identification technique.
\newblock In \emph{36th Annual Meeting of the Association for Computational
  Linguistics and 17th International Conference on Computational Linguistics,
  Volume 1}, pages 206--210.

\bibitem[{Calhoun et~al.(2010)Calhoun, Carletta, Brenier, Mayo, Jurafsky,
  Steedman, and Beaver}]{calhoun2010nxt}
Sasha Calhoun, Jean Carletta, Jason~M Brenier, Neil Mayo, Dan Jurafsky, Mark
  Steedman, and David Beaver. 2010.
\newblock The nxt-format switchboard corpus: a rich resource for investigating
  the syntax, semantics, pragmatics and prosody of dialogue.
\newblock \emph{Language resources and evaluation}, 44(4):387--419.

\bibitem[{Carlini et~al.(2014)Carlini, Codina-Filba, and
  Wanner}]{carlini2014improving}
Roberto Carlini, Joan Codina-Filba, and Leo Wanner. 2014.
\newblock Improving collocation correction by ranking suggestions using
  linguistic knowledge.
\newblock In \emph{Proceedings of the third workshop on NLP for
  computer-assisted language learning}, pages 1--12.

\bibitem[{Cazden(1988)}]{cazden1988classroom}
Courtney~B Cazden. 1988.
\newblock \emph{Classroom discourse: The language of teaching and learning.}
\newblock ERIC.

\bibitem[{Chapin(2003)}]{Chapin2003ClassroomDU}
Suzanne~H. Chapin. 2003.
\newblock Classroom discussions: Using math talk to help students learn, grades
  1-6.

\bibitem[{Chapin et~al.(2009)Chapin, O'Connor, O'Connor, and
  Anderson}]{chapin2009classroom}
Suzanne~H Chapin, Catherine O'Connor, Mary~Catherine O'Connor, and
  Nancy~Canavan Anderson. 2009.
\newblock \emph{Classroom discussions: Using math talk to help students learn,
  Grades K-6}.
\newblock Math Solutions.

\bibitem[{Chi and Wylie(2014)}]{chi2014icap}
Michelene~TH Chi and Ruth Wylie. 2014.
\newblock The icap framework: Linking cognitive engagement to active learning
  outcomes.
\newblock \emph{Educational psychologist}, 49(4):219--243.

\bibitem[{Cho et~al.(2014)Cho, van Merri{\"e}nboer, Gulcehre, Bahdanau,
  Bougares, Schwenk, and Bengio}]{cho-etal-2014-learning}
Kyunghyun Cho, Bart van Merri{\"e}nboer, Caglar Gulcehre, Dzmitry Bahdanau,
  Fethi Bougares, Holger Schwenk, and Yoshua Bengio. 2014.
\newblock \href {https://doi.org/10.3115/v1/D14-1179} {Learning phrase
  representations using {RNN} encoder{--}decoder for statistical machine
  translation}.
\newblock In \emph{Proceedings of the 2014 Conference on Empirical Methods in
  Natural Language Processing ({EMNLP})}, pages 1724--1734, Doha, Qatar.
  Association for Computational Linguistics.

\bibitem[{Chollampatt and Ng(2018)}]{chollampatt2018multilayer}
Shamil Chollampatt and Hwee~Tou Ng. 2018.
\newblock A multilayer convolutional encoder-decoder neural network for
  grammatical error correction.
\newblock In \emph{Proceedings of the AAAI Conference on Artificial
  Intelligence}, volume~32.

\bibitem[{Chukharev-Hudilainen and Saricaoglu(2016)}]{chukharev2016causal}
Evgeny Chukharev-Hudilainen and Aysel Saricaoglu. 2016.
\newblock Causal discourse analyzer: Improving automated feedback on academic
  esl writing.
\newblock \emph{Computer Assisted Language Learning}, 29(3):494--516.

\bibitem[{Dahlmeier and Ng(2011)}]{dahlmeier2011correcting}
Daniel Dahlmeier and Hwee~Tou Ng. 2011.
\newblock Correcting semantic collocation errors with l1-induced paraphrases.
\newblock In \emph{Proceedings of the 2011 conference on empirical methods in
  natural language processing}, pages 107--117.

\bibitem[{Donnelly et~al.(2016)Donnelly, Blanchard, Samei, Olney, Sun, Ward,
  Kelly, Nystran, and D'Mello}]{donnelly}
Patrick~J. Donnelly, Nathan Blanchard, Borhan Samei, Andrew~M. Olney, Xiaoyi
  Sun, Brooke Ward, Sean Kelly, Martin Nystran, and Sidney~K. D'Mello. 2016.
\newblock \href {https://doi.org/10.1145/2930238.2930250} {Automatic teacher
  modeling from live classroom audio}.
\newblock In \emph{Proceedings of the 2016 Conference on User Modeling
  Adaptation and Personalization}, UMAP '16, page 45–53, New York, NY, USA.
  Association for Computing Machinery.

\bibitem[{Dyke et~al.(2013)Dyke, Howley, Adamson, Kumar, and
  Ros{\'e}}]{dyke2013towards}
Gregory Dyke, Iris Howley, David Adamson, Rohit Kumar, and Carolyn~Penstein
  Ros{\'e}. 2013.
\newblock Towards academically productive talk supported by conversational
  agents.
\newblock In \emph{Productive multivocality in the analysis of group
  interactions}, pages 459--476. Springer.

\bibitem[{Farag et~al.(2018)Farag, Yannakoudakis, and
  Briscoe}]{farag2018neural}
Youmna Farag, Helen Yannakoudakis, and Ted Briscoe. 2018.
\newblock Neural automated essay scoring and coherence modeling for
  adversarially crafted input.
\newblock \emph{arXiv preprint arXiv:1804.06898}.

\bibitem[{Goldenberg(1992)}]{goldenberg1992instructional}
Claude Goldenberg. 1992.
\newblock Instructional conversations: Promoting comprehension through
  discussion.
\newblock \emph{The Reading Teacher}, 46(4):316--326.

\bibitem[{Khanpour et~al.(2016)Khanpour, Guntakandla, and
  Nielsen}]{khanpour-etal-2016-dialogue}
Hamed Khanpour, Nishitha Guntakandla, and Rodney Nielsen. 2016.
\newblock \href {https://www.aclweb.org/anthology/C16-1189} {Dialogue act
  classification in domain-independent conversations using a deep recurrent
  neural network}.
\newblock In \emph{Proceedings of {COLING} 2016, the 26th International
  Conference on Computational Linguistics: Technical Papers}, pages 2012--2021,
  Osaka, Japan. The COLING 2016 Organizing Committee.

\bibitem[{Kingma and Ba(2014)}]{kingma2014adam}
Diederik~P Kingma and Jimmy Ba. 2014.
\newblock Adam: A method for stochastic optimization.
\newblock \emph{arXiv preprint arXiv:1412.6980}.

\bibitem[{Litman(2016)}]{litman2016natural}
Diane Litman. 2016.
\newblock Natural language processing for enhancing teaching and learning.
\newblock In \emph{Proceedings of the AAAI Conference on Artificial
  Intelligence}, volume~30.

\bibitem[{Liu et~al.(2019)Liu, Ott, Goyal, Du, Joshi, Chen, Levy, Lewis,
  Zettlemoyer, and Stoyanov}]{Liu2019RoBERTaAR}
Y.~Liu, Myle Ott, Naman Goyal, Jingfei Du, Mandar Joshi, Danqi Chen, Omer Levy,
  M.~Lewis, Luke Zettlemoyer, and Veselin Stoyanov. 2019.
\newblock Roberta: A robustly optimized bert pretraining approach.
\newblock \emph{ArXiv}, abs/1907.11692.

\bibitem[{Loper and Bird(2002)}]{nltk}
Edward Loper and Steven Bird. 2002.
\newblock \href {https://doi.org/10.3115/1118108.1118117} {Nltk: The natural
  language toolkit}.
\newblock In \emph{Proceedings of the ACL-02 Workshop on Effective Tools and
  Methodologies for Teaching Natural Language Processing and Computational
  Linguistics - Volume 1}, ETMTNLP '02, page 63–70, USA. Association for
  Computational Linguistics.

\bibitem[{McNamara(2011)}]{mcnamara2011measuring}
Danielle~S McNamara. 2011.
\newblock Measuring deep, reflective comprehension and learning strategies:
  challenges and successes.
\newblock \emph{Metacognition and Learning}, 6(2):195--203.

\bibitem[{McNamara et~al.(2013)McNamara, Crossley, and
  Roscoe}]{mcnamara2013natural}
Danielle~S McNamara, Scott~A Crossley, and Rod Roscoe. 2013.
\newblock Natural language processing in an intelligent writing strategy
  tutoring system.
\newblock \emph{Behavior research methods}, 45(2):499--515.

\bibitem[{Michaels and O’Connor(2015)}]{michaels2015conceptualizing}
Sarah Michaels and Catherine O’Connor. 2015.
\newblock Conceptualizing talk moves as tools: Professional development
  approaches for academically productive discussion.
\newblock \emph{Socializing intelligence through talk and dialogue}, pages
  347--362.

\bibitem[{Michaels et~al.(2008)Michaels, O’Connor, and
  Resnick}]{michaels2008deliberative}
Sarah Michaels, Catherine O’Connor, and Lauren~B Resnick. 2008.
\newblock Deliberative discourse idealized and realized: Accountable talk in
  the classroom and in civic life.
\newblock \emph{Studies in philosophy and education}, 27(4):283--297.

\bibitem[{Ott et~al.(2019)Ott, Edunov, Baevski, Fan, Gross, Ng, Grangier, and
  Auli}]{ott-etal-2019-fairseq}
Myle Ott, Sergey Edunov, Alexei Baevski, Angela Fan, Sam Gross, Nathan Ng,
  David Grangier, and Michael Auli. 2019.
\newblock \href {https://doi.org/10.18653/v1/N19-4009} {fairseq: A fast,
  extensible toolkit for sequence modeling}.
\newblock In \emph{Proceedings of the 2019 Conference of the North {A}merican
  Chapter of the Association for Computational Linguistics (Demonstrations)},
  pages 48--53, Minneapolis, Minnesota. Association for Computational
  Linguistics.

\bibitem[{O’Connor and Michaels(2019)}]{o2019supporting}
Catherine O’Connor and Sarah Michaels. 2019.
\newblock Supporting teachers in taking up productive talk moves: The long road
  to professional learning at scale.
\newblock \emph{International Journal of Educational Research}, 97:166--175.

\bibitem[{Paszke et~al.(2019)Paszke, Gross, Massa, Lerer, Bradbury, Chanan,
  Killeen, Lin, Gimelshein, Antiga, Desmaison, Kopf, Yang, DeVito, Raison,
  Tejani, Chilamkurthy, Steiner, Fang, Bai, and Chintala}]{NEURIPS2019_9015}
Adam Paszke, Sam Gross, Francisco Massa, Adam Lerer, James Bradbury, Gregory
  Chanan, Trevor Killeen, Zeming Lin, Natalia Gimelshein, Luca Antiga, Alban
  Desmaison, Andreas Kopf, Edward Yang, Zachary DeVito, Martin Raison, Alykhan
  Tejani, Sasank Chilamkurthy, Benoit Steiner, Lu~Fang, Junjie Bai, and Soumith
  Chintala. 2019.
\newblock \href
  {http://papers.neurips.cc/paper/9015-pytorch-an-imperative-style-high-performance-deep-learning-library.pdf}
  {Pytorch: An imperative style, high-performance deep learning library}.
\newblock In H.~Wallach, H.~Larochelle, A.~Beygelzimer, F.~d\textquotesingle
  Alch\'{e}-Buc, E.~Fox, and R.~Garnett, editors, \emph{Advances in Neural
  Information Processing Systems 32}, pages 8024--8035. Curran Associates, Inc.

\bibitem[{Roller et~al.(2020)Roller, Dinan, Goyal, Ju, Williamson, Liu, Xu,
  Ott, Shuster, Smith, Boureau, and Weston}]{roller2020recipes}
Stephen Roller, Emily Dinan, Naman Goyal, Da~Ju, Mary Williamson, Yinhan Liu,
  Jing Xu, Myle Ott, Kurt Shuster, Eric~M. Smith, Y-Lan Boureau, and Jason
  Weston. 2020.
\newblock \href {http://arxiv.org/abs/2004.13637} {Recipes for building an
  open-domain chatbot}.

\bibitem[{Sennrich et~al.(2016)Sennrich, Haddow, and Birch}]{sennrich-tokenize}
Rico Sennrich, Barry Haddow, and Alexandra Birch. 2016.
\newblock \href {https://doi.org/10.18653/v1/P16-1162} {Neural machine
  translation of rare words with subword units}.
\newblock In \emph{Proceedings of the 54th Annual Meeting of the Association
  for Computational Linguistics (Volume 1: Long Papers)}, pages 1715--1725,
  Berlin, Germany. Association for Computational Linguistics.

\bibitem[{Stolcke et~al.(2000)Stolcke, Ries, Coccaro, Shriberg, Bates,
  Jurafsky, Taylor, Martin, Ess-Dykema, and Meteer}]{stolcke2000dialogue}
Andreas Stolcke, Klaus Ries, Noah Coccaro, Elizabeth Shriberg, Rebecca Bates,
  Daniel Jurafsky, Paul Taylor, Rachel Martin, Carol~Van Ess-Dykema, and Marie
  Meteer. 2000.
\newblock Dialogue act modeling for automatic tagging and recognition of
  conversational speech.
\newblock \emph{Computational linguistics}, 26(3):339--373.

\bibitem[{Suresh et~al.(2019)Suresh, Sumner, Jacobs, Foland, and
  Ward}]{suresh2019automating}
Abhijit Suresh, Tamara Sumner, Jennifer Jacobs, Bill Foland, and Wayne Ward.
  2019.
\newblock Automating analysis and feedback to improve mathematics teachers’
  classroom discourse.
\newblock In \emph{Proceedings of the AAAI Conference on Artificial
  Intelligence}, volume~33, pages 9721--9728.

\bibitem[{Tanaka et~al.(2019)Tanaka, Takayama, and
  Arase}]{tanaka-etal-2019-dialogue}
Koji Tanaka, Junya Takayama, and Yuki Arase. 2019.
\newblock \href {https://doi.org/10.18653/v1/P19-2027} {Dialogue-act prediction
  of future responses based on conversation history}.
\newblock In \emph{Proceedings of the 57th Annual Meeting of the Association
  for Computational Linguistics: Student Research Workshop}, pages 197--202,
  Florence, Italy. Association for Computational Linguistics.

\bibitem[{Tegos et~al.(2015)Tegos, Demetriadis, and
  Karakostas}]{tegos2015promoting}
Stergios Tegos, Stavros Demetriadis, and Anastasios Karakostas. 2015.
\newblock Promoting academically productive talk with conversational agent
  interventions in collaborative learning settings.
\newblock \emph{Computers \& Education}, 87:309--325.

\bibitem[{Tegos et~al.(2016)Tegos, Demetriadis, Papadopoulos, and
  Weinberger}]{tegos2016conversational}
Stergios Tegos, Stavros Demetriadis, Pantelis~M Papadopoulos, and Armin
  Weinberger. 2016.
\newblock Conversational agents for academically productive talk: A comparison
  of directed and undirected agent interventions.
\newblock \emph{International Journal of Computer-Supported Collaborative
  Learning}, 11(4):417--440.

\bibitem[{Tetreault et~al.(2014)Tetreault, Chodorow, and
  Madnani}]{tetreault2014bucking}
Joel Tetreault, Martin Chodorow, and Nitin Madnani. 2014.
\newblock Bucking the trend: improved evaluation and annotation practices for
  esl error detection systems.
\newblock \emph{Language Resources and Evaluation}, 48(1):5--31.

\bibitem[{Wang et~al.(2019)Wang, Pruksachatkun, Nangia, Singh, Michael, Hill,
  Levy, and Bowman}]{superglue}
Alex Wang, Yada Pruksachatkun, Nikita Nangia, Amanpreet Singh, Julian Michael,
  Felix Hill, Omer Levy, and Samuel Bowman. 2019.
\newblock \href
  {https://proceedings.neurips.cc/paper/2019/file/4496bf24afe7fab6f046bf4923da8de6-Paper.pdf}
  {Superglue: A stickier benchmark for general-purpose language understanding
  systems}.
\newblock In \emph{Advances in Neural Information Processing Systems},
  volume~32. Curran Associates, Inc.

\bibitem[{Wu et~al.(2020)Wu, Hoi, Socher, and Xiong}]{wu-etal-2020-tod}
Chien-Sheng Wu, Steven~C.H. Hoi, Richard Socher, and Caiming Xiong. 2020.
\newblock \href {https://doi.org/10.18653/v1/2020.emnlp-main.66} {{TOD}-{BERT}:
  Pre-trained natural language understanding for task-oriented dialogue}.
\newblock In \emph{Proceedings of the 2020 Conference on Empirical Methods in
  Natural Language Processing (EMNLP)}, pages 917--929, Online. Association for
  Computational Linguistics.

\bibitem[{Yu and Yu(2019)}]{yudialog}
Dian Yu and Zhou Yu. 2019.
\newblock \href {http://arxiv.org/abs/1908.10023} {{MIDAS:} {A} dialog act
  annotation scheme for open domain human machine spoken conversations}.
\newblock \emph{CoRR}, abs/1908.10023.

\bibitem[{Zhang et~al.(2020)Zhang, Sun, Galley, Chen, Brockett, Gao, Gao, Liu,
  and Dolan}]{zhang-etal-2020-dialogpt}
Yizhe Zhang, Siqi Sun, Michel Galley, Yen-Chun Chen, Chris Brockett, Xiang Gao,
  Jianfeng Gao, Jingjing Liu, and Bill Dolan. 2020.
\newblock \href {https://doi.org/10.18653/v1/2020.acl-demos.30} {{DIALOGPT} :
  Large-scale generative pre-training for conversational response generation}.
\newblock In \emph{Proceedings of the 58th Annual Meeting of the Association
  for Computational Linguistics: System Demonstrations}, pages 270--278,
  Online. Association for Computational Linguistics.

\end{thebibliography}

\appendix
\clearpage

\section{Tuning of the Window Size}
\begin{table}[h!]
    \centering
    \begin{tabular}{|p{2.5cm}|p{0.9cm}|p{0.9cm}|p{0.9cm}|p{0.9cm}|}
 \hline
 Configuration & Prec. & Recall & F1 & Acc. \\ [0.5ex]
 \hline\hline
 No weighting, window 5 & 32.31 & 25.67 & 26.94 & 63.44 \\
 \hline
 Class weighting, window 5 & 34.84 & 29.39 & 29.83 & 62.24 \\
 \hline
 Class weighting, window 1 & 31.17 & 29.18 & 28.64 & 62.61 \\
 \hline
 Class weighting, window 2 & 34.11 & 29.44 & 27.62 & 62.85 \\
 \hline
 Class weighting, window 3 & 29.83 & 29.29 & 29.09 & 58.07 \\
 \hline
 Class weighting, window 4 & 29.72 & 28.57 & 28.51 & 61.65 \\
 \hline
 Class weighting, window 6 & 32.55 & 30.56 & 28.95 & 63.76 \\
 \hline
 Class weighting, window 7 & 33.21 & 28.64 & 29.05 & 58.27 \\
 \hline
\end{tabular}
\caption{Tuning experiments on the development set}
\label{tab:tuning}
\end{table}

\section{Training and Hyperparameters} 
3-E is implemented in PyTorch \cite{NEURIPS2019_9015}. 
For training, we use an Adam optimizer \cite{kingma2014adam} with an initial learning rate of $1e^{-4}$ and train for 30 epochs. The utterance encoder has an embedding size of 256, and a hidden layer size of 512. The talk move encoder has an embedding size of 32 and a hidden layer size of 64. The dialogue encoder has a hidden layer size of 1025. Finally, the feedforward layer uses a hidden layer size of 32. We do not use pretrained word embeddings in the 3-E model.

For the 3-E-RoBERTa model, we use the pretrained parameters of the Fairseq library's implementation of RoBERTa \cite{ott-etal-2019-fairseq}, and use representations with dimension 1024. For 3-E-RoBERTa, we further add dropout with a probability of 0.4 in two places to avoid over-fitting: on the layer where the two utterance representations are concatenated (c.f. Equation \ref{eq:roberta}), and the layer where the utterance history and talk move history are concatenated (c.f. Equation \ref{eq:concattm}). 

For 3-E-TOD-BERT, we use the pretrained model provided by \citet{wu-etal-2020-tod} trained jointly on masked language modeling and response contrastive loss. We additionally add a dropout of 0.2 at the layer where the utterance representation is concatenated with talk move history (c.f. Equation \ref{eq:tod}).

The baseline models TM-only-w and TM-only-z are trained for 30 epochs using the same hyperparameters, with a batch size of 256 and a learning rate of 1e-4.

\end{document}